# COMBINING THE ANALYTICAL HIERARCHY PROCESS AND THE GENETIC ALGORITHM TO SOLVE THE TIMETABLE PROBLEM

Ihab Sbeity, Mohamed Dbouk and Habib Kobeissi

Department of Computer Sciences, Faculty of Sciences, Section I, Lebanese University, LEBANON

## ABSTRACT

*The main problems of school course timetabling are time, curriculum, and classrooms. In addition there are other problems that vary from one institution to another. This paper is intended to solve the problem of satisfying the teachers' preferred schedule in a way that regards the importance of the teacher to the supervising institute, i.e. his score according to some criteria. Genetic algorithm (GA) has been presented as an elegant method in solving timetable problem (TTP) in order to produce solutions with no conflict. In this paper, we consider the analytic hierarchy process (AHP) to efficiently obtain a score for each teacher, and consequently produce a GA-based TTP solution that satisfies most of the teachers' preferences.*

## 1. INTRODUCTION

The timetable problem (TTP) is the process of scheduling a sequence of courses between teachers, students and rooms, to satisfy a set of various constraints. These constraints differ based on the institutions involved, e.g. schools, universities, and so on.

In some institutions, such as *Lebanese* elementary schools and high-schools, an important constraint to consider is the teacher preferences that may be related to some criteria, such as the teacher age, his address, his contract-type, etc. For example, a teacher with seniority should have the opportunity to choose his course schedule. Or, a part-time teacher has the priority to select his preferred schedule over a full-time teacher, and so on. Some of these criteria may be a reflection of the social life and relationships in the urban Lebanese institutions. Note that, these criteria may be conflicting so that the ranking of teachers accordingly becomes not obvious. This is, for example, the case when comparing a senior full-time teacher with a younger part-time teacher.

The analytic hierarchy process (AHP), developed by Saaty [1], is an effective tool for dealing with such complex decision making processes; therefore building a ranking relationship between teachers, based on a series of pairwise comparisons. In addition, the AHP incorporates a useful technique for checking the consistency of the decision evaluation, hence reducing the bias in the decision making process.

On the other hand, the manual solution of TTP is still used nowadays, for reasons related, not only to the lack of budget needed to buy a dedicated TTP solver such that aSc [2, 3] and Prime [4], but also to the inconsistency of such tools with the Lebanese teachers' criteria. With the majority of existing tools, the person responsible in creating the schedule must manually specify the teachers' priorities with their -inconsistent-preferences, and then the tool will generate the timetable schedule. However, the teachers' priorities are not always evident to be significantly measured. In addition, as TTP is an NP-complete problem, its manual solution is time and effort-consuming. The problem would be presented by a great number of variables which make it intractable. That was the motivation to approach it by means of Genetic Algorithm (GA) [5].

In this paper, we present a new approach (AHP/GA) that combines AHP and GA to create a time table schedule that matches most of the teachers' preferences along with the institution. Using AHP, we will develop a teachers ranking by providing a score for each teacher, and then, a *satisfaction function* will be incorporated to the GA to produce the schedule that matches as best the AHP ranking, and satisfies of most the teachers' preferences. An informal description of the approach is presented basing on a simple example. The formal description and the implementation of the method would be our future challenge.

The rest of this paper is structured as follows: Section 2 presents a background of works related to the utilization of GA to solve TTP. Section 3 describes the steps followed by AHP/GA to produce a time table schedule taking into consideration the teachers' preferences. We also present a simple case study to be an application of the procedure. In section 3, AHP is applied to a set of teachers and their score calculation procedure is presented. Section 4 describes the result of the combination of AHP and GA to generate the time table schedule. Section 5 concludes our paper and describes our ongoing works.

## 2. BACKGROUND

### 2.1. Genetic algorithm

A genetic algorithm is a type of searching algorithm. It searches a solution space for an optimal solution to a problem. The key characteristic of the genetic algorithm is how the searching is done. The algorithm creates a "population" of possible solutions to the problem and lets them "evolve" over multiple generations to find better and better solutions. The population is the collection of candidate solutions that we are considering during the course of the algorithm. Over the generations of the algorithm, new members are "born" into the population, while others "die" out of the population. A single solution in the population is referred to as an individual. *A fitness function* of an individual presents a measure of how "good" the solution represented by the individual is. The better the solution, the higher the fitness – obviously, this is dependent on the problem to be solved.

The selection process is analogous to the survival of the fittest in the natural world. Individuals are selected for "breeding" (or cross-over) based upon their fitness values –the fitter the individual, the more likely that individual will be able to reproduce. The cross-over occurs by mingling the two solutions together to produce two new individuals. During each generation, there is a small chance for each individual to mutate, which will change the individual in some small way [6]. For further information about GA, refer to [7].

### 2.2. Related works

Various types of timetable problems appear with regards to educational institutions and depend on the elements to be scheduled, i.e. exams or regular courses. An excellent survey of exam timetabling techniques is presented in Qu et al. [8]. The examination timetabling is similar in most institutions and consists of scheduling the exams for a set of courses, over a limited time period, while avoiding the overlapping of the exams for each student, as well as seeking the largest spread over the examination period [9].

However, the course timetable differs regarding to the institution, i.e. universities, schools. A course timetable scheduling problem basically deals with effective distribution of five kinds of entities (or resources): Rooms, Courses, Teachers, Students, and Time (days and hours).

In university timetables, the problem presents a large number of constraints to be satisfied. These constraints typically cover the time conflict, the room conflict, the course conflict, and other criteria related to teachers and students preferences. Commonly, they are divided into two types: hard constraints and soft constraints [10]. Hard constraints are those that cannot be violated, such as: "no more than one course is allowed at a timeslot in each room." Soft constraints may be violated, but the purpose lies in minimizing their violation. An example of soft constraints in a university timetable is: "a student can only attend one course per day". A wave of works had proposed GA-based solutions for the university timetable problem; we underline the works of [10, 11, 12].

The schools TTP may be defined as a subset of the universities TTP, as in university TTP the number of set of constraints is larger. Several works has addressed the school TTP and presented different approaches to solve the problem; we highlight the works of [8, 13, 14]. Each of these works proposes a different utilization and definition of GA parameters (chromosomes description, mutation and crossover definition, etc). Other works apply a parallel implementation of GA to solve the school TTP [15]. However, addressing the teachers' preferences in the way presented by this paper has never been previously carried out, and it would be our contribution.

## 3. THE AHP/GA PROCESS AND PROBLEM DESCRIPTION

A common classic definition of the GA fitness function presented by previous works [13, 14, 15, 16] takes into consideration the time conflict. Hence, our AHP/GA approach assumes that the time conflict is already solved by using one of the fitness functions previously mentioned. Our concern is to consider the teachers' preferences such that a feasible solution is the one that satisfies the most of these preferences. The AHP/GA process is described by figure 1.

First, AHP is used to calculate the score of each teacher arising from a set of information given according to the institution valuable criteria. These scores are considered as one of the GA input. The fitness function of the GA is practically composed of two functions: the conflict function and the *satisfaction function*. A feasible solution should be validated by both functions; however, we omit the process used by the conflict function as it may be the same as in one of previous mentioned works. The satisfaction function, having the teachers' scores as parameters, checks if the GA produced solutions satisfy the teachers' preferences given as input.

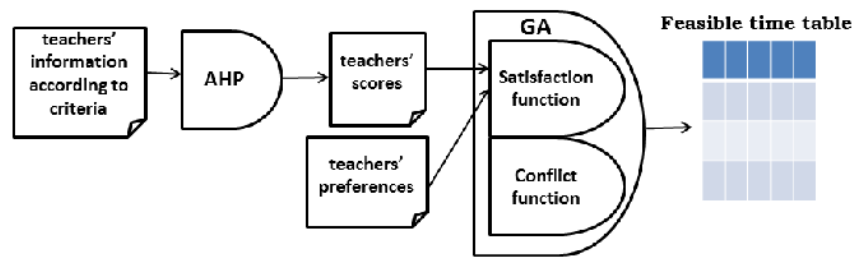

Figure 1: the AHP/GA process

In the rest of the paper, the AHP/GA process is applied on a simple case study described below.

**Description of a school TTP**

In this paper, we only present an informal description of our approach. Having a purpose to simplify the problem description and the presentation of its solution later on, we consider the following simple TTP model:

- The school is open only 3 days a week, e.g. D1, D2, and D3.
- Daily, 4 sessions are only scheduled on 4 time slots, each two are separated by a break session (BS).
- The school provides 2 classes: C1 and C2. Thus, each class has (3×4) 12 sessions per week, and a total of (12×2) 24 sessions are achieved weekly by all the school's teachers.
- The school has 6 teachers. The distribution of the set of sessions over the teachers is given in the following table (Table 1):

Table 1: Distribution of sessions over the set of teachers

| Teacher | C1 | C2 | Total |
|---------|----|----|-------|
| T1 | 3 | 4 | 7 |
| T2 | 2 | 2 | 4 |
| T3 | 2 | 1 | 3 |
| T4 | 2 | 3 | 5 |
| T5 | 1 | 1 | 2 |
| T6 | 2 | 1 | 3 |
| **Total** | 12 | 12 | **24** |

Note that in our study, we do not take into consideration the course name (math, sciences, literature, etc..); therefore there is no need to specify the course given in each session, as each course is assigned to only one teacher. Accordingly, only the teacher's schedule should be implemented without taking care of the course name.

Recalling that our purpose is to simplify the presentation of the approach, the above model is a simple description of reality. A standard Lebanese school actually provides nine levels (classes) that may have one or more sections each depending on the number of students tolerable in the level (in case of two or more sections by level, we may consider each section as a new separate class). The school is usually open five days a week, and offers daily six sections per class. Thus, at least a total of (9×5×6) 270 sessions should be scheduled over the nine classes and the set of teachers.

Back to our simple model, table 2 presents an empty time slot table that should be assigned to teachers.

Table 2: Empty time table

| C1 | | | C2 | | |
|----|----|----|----|----|----|
| D1 | D2 | D3 | D1 | D2 | D3 |
| | | | | | |
| | | | | | |
| | | | | | |
| | | | | | |

As in previous related works, here again, two kinds of constraints should be respected in the time table schedule, hard and soft. We restrict our study on necessary constraints. The Other constraints (such as "for a class, two consecutive sessions of the same teacher is not acceptable) are not considered, but their integration to our approach seems to be promising.

- **Hard constraints**:
  1. There should not be time conflict in the teacher schedule, i.e. a teacher should not have two or more sessions in parallel in the same time slot.
  2. Each teacher should be assigned the specified number of session by class.

- **Soft constraint**: only one soft constraint has to be underlined, and it is related to the teacher preferences. A teacher would specify a proffered schedule. The goal is to satisfy at most the teachers' preferences. When all the teachers' preferences are satisfied, a *perfect* time table schedule is obtained.

Table 3: Preferred schedule of teachers

| T1 | | | T2 | | | T3 | | |
|---|---|---|---|---|---|---|---|---|
| D1 | D2 | D3 | D1 | D2 | D3 | D1 | D2 | D3 |
| 1 | | 1 | | 1 | | 1 | | 1 |
| 1 | | 1 | | 1 | | | | 1 |
| 1 | 1 | | 1 | | 1 | | 1 | 1 |
| 1 | 1 | | 1 | | 1 | | 1 | |
| **T4** | | | **T5** | | | **T6** | | |
| D1 | D2 | D3 | D1 | D2 | D3 | D1 | D2 | D3 |
| 1 | 1 | 1 | | 1 | | 1 | | 1 |
| 1 | 1 | 1 | | 1 | | 1 | | 1 |
| 1 | 1 | 1 | | 1 | | | 1 | 1 |
| 1 | 1 | 1 | 1 | 1 | 1 | | 1 | |

As it is mentioned above, the described hard constraints are already addressed by previous listed works. In our paper, we suppose that the GA is able to produce a time table without time conflict, and that each teacher is assigned an exact number of sessions as required for them. Respecting the soft constraint is the key issue that we intend to solve by satisfying the teachers' preferences. In order to do that, we assume that each teacher should provide his preferred schedule to the institution. Our combined (AHP/GA) algorithm will then generate the time table that satisfies the teachers' preferred schedule according to a *satisfaction threshold (ST)* fixed by the institution. The *satisfaction function (SF)* will be applied to each possible solution generated by (AHP/GA). Hence, a solution is feasible when its SF is greater than ST.

For our simple case, table 3 presents the preferred schedule of six teachers. The value '1' means that the specified session is preferred by the teacher. For example, T1 has no problem with his schedule in D1. However, he prefers to give his sessions after the break session in D2, and before the break session in D3.

In the next section, we show how to generate a score for each teacher using AHP. These scores are the crucial parameters of the satisfaction function.

## 4. THE ANALYTICAL HIERARCHY PROCESS

The AHP is a simple decision-making tool that deals with complex multi-attribute problems which has been developed by Thomas Saaty in the (1980's). It is widely known as a method for ranking decision alternatives and selecting the best one when the decision maker has multiple objectives, or criteria. The selection process is based on the calculation of scores for alternatives. The key point in the AHP method is the pairwise comparison used to calculate the relative weights of criteria, and consequently to develop an overall ranking of alternatives. In order to help the pairwise comparison, Saaty created a nine-point scale of importance between two elements. Here, we are interested in only five scales. The suggested numbers to express degrees of preference between the two elements is shown in table 4.

Table 4: Standard scale used in AHP

| Preference level | Numerical value |
|---|---|
| Equally preferred | 1 |
| Moderately preferred | 3 |
| Strongly preferred | 5 |
| Very Strongly preferred | 7 |
| Extremely preferred | 9 |

In our approach, AHP is used in order to generate a score $S_i$ for each teacher $T_i$. For this goal, we present the steps followed in AHP method by applying to our simplified TTP example.

In our case study, we are interested in ranking the list of teachers (alternatives) according to a list of criteria – supposed - defined by the school and that can differ from an institution to another. Once again, to simplify the presentation of the approach, we consider four criteria: the teacher's age, the teacher's contract-type (part/full timer), the teacher's gender (male/female), and the teacher's teaching load. Other criteria related to teachers such as health conditions, marital status, distance between the living address and the school, may also be significant. *The teaching load $L_i$ of each teacher $T_i$ is also an essential data needed later on in the (AHP/GA) approach, even if it is not considered as the teacher criteria.* Table 5 presents the list of information about the 6 teachers of our case study, according to the 4 criteria.

For the set of criteria, the school should specify the importance of a criterion compared to another criterion. The order of importance is translated to a value using the AHP scale of table 4.

Table 5:  Information about teachers

| Teacher | Age | Gender | Contract | Load |
|---|---|---|---|---|
| T1 | 42 | M | Full | 7 |
| T2 | 33 | F | Full | 4 |
| T3 | 25 | F | Part | 3 |
| T4 | 24 | F | Part | 5 |
| T5 | 63 | M | Full | 2 |
| T6 | 43 | F | Full | 3 |

We consider that the contract type and the teacher load are the most important factors that give the teacher the superiority to choose his/her preferred schedule. A part-timer teacher has primacy over a full-timer. Also, the teacher with greater load should be prioritized over a teacher with a smaller load.

The teacher's age comes next in the importance of criteria. Ages are divided into 5 intervals: I1 = [21 - 30], I2 = [31 - 40], I3 = [41 - 50], I4 = [50 - 60] and I5 = [60 – and up]. Intervals of older ages become more important in the preference level, e.g. older teachers are given the priority to choose their course schedule over younger teachers. Teachers of the same age interval have the same preference value.

The teacher's gender has the least importance in the preference order. We assume that a "female" teacher has the priority in the preference level.

According to the AHP scale of table 4, and the relative importance of criteria described above, the pairwise comparison rating for each criterion is given in table 6.

Table 6: Pairwise comparison for the four criteria

| Age | | | | | | | | Gender | | | | | | |
|---|---|---|---|---|---|---|---|---|---|---|---|---|---|---|
| Teacher | T1 | T2 | T3 | T4 | T5 | T6 | | Teacher | T1 | T2 | T3 | T4 | T5 | T6 |
| T1 | 1 | 3 | 5 | 5 | 1/5 | 1 | | T1 | 1 | 1/3 | 1/3 | 1/3 | 1 | 1/3 |
| T2 | 1/3 | 1 | 3 | 3 | 1/7 | 1/3 | | T2 | 3 | 1 | 1 | 1 | 3 | 1 |
| T3 | 1/5 | 1/3 | 1 | 1 | 1/9 | 1/5 | | T3 | 3 | 1 | 1 | 1 | 3 | 1 |
| T4 | 1/5 | 1/3 | 1 | 1 | 1/9 | 1/5 | | T4 | 3 | 1 | 1 | 1 | 3 | 1 |
| T5 | 5 | 7 | 9 | 9 | 1 | 5 | | T5 | 1 | 1/3 | 1/3 | 1/3 | 1 | 1/3 |
| T6 | 1 | 3 | 5 | 5 | 1/5 | 1 | | T6 | 3 | 1 | 1 | 1 | 3 | 1 |

| Load | | | | | | | | Contract type | | | | | | |
|---|---|---|---|---|---|---|---|---|---|---|---|---|---|---|
| Teacher | T1 | T2 | T3 | T4 | T5 | T6 | | Teacher | T1 | T2 | T3 | T4 | T5 | T6 |
| T1 | 1 | 3 | 5 | 3 | 5 | 5 | | T1 | 1 | 1 | 1/5 | 1/5 | 1 | 1 |
| T2 | 1/3 | 1 | 3 | 1 | 3 | 3 | | T2 | 1 | 1 | 1/5 | 1/5 | 1 | 1 |
| T3 | 1/5 | 1/3 | 1 | 1/3 | 1 | 1 | | T3 | 5 | 5 | 1 | 1 | 5 | 5 |
| T4 | 1/3 | 1 | 3 | 1 | 3 | 3 | | T4 | 5 | 5 | 1 | 1 | 5 | 5 |
| T5 | 1/5 | 1/3 | 1 | 1/3 | 1 | 1 | | T5 | 1 | 1 | 1/5 | 1/5 | 1 | 1 |
| T6 | 1/5 | 1/3 | 1 | 1/3 | 1 | 1 | | T6 | 1 | 1 | 1/5 | 1/5 | 1 | 1 |

As per the criterion "AGE", we consider that the ascending exploration of intervals correspond to a same exploration in the preference scale. Thus, the age of the interval I3 is "strongly preferred" to the age of I2, and "moderately preferred" to the age of I1, and so on. As an example, the pairwise comparison of T1 to T3 gives the value 5 meaning that the age of T1 (42) is strongly preferred to the age of T3 (24).

The pairwise comparison according to the criterion "gender" respects the fact that a female is moderately preferred to a male teacher. The pairwise comparison of the criterion "contract-type" is achieved according to the fact that a part-timer is strongly preferred to a full-timer. Finally, we applied the pairwise comparison of the "load" criteria, again, basing on a division of the load value into intervals [0-2[, [2-4[, [4-6[, [6-..[, and the same strategy used for the criterion "age" is applied.

Note that for all criteria, any teacher compared to himself is equally preferred, and that if $T_i$ is compared to $T_j$ for a criterion and the preference value is x, then the preference value of comparing $T_j$ to $T_i$ is 1/x.

The second step in the AHP process is the generation of the preference vectors of each criterion. In table 7, we present how to calculate the preference vector of the criterion "gender".

Table 7: Calculation of the preference vector for criterion "Gender"

| Gender | | | | | | | |
|---|---|---|---|---|---|---|---|
| Teacher | T1 | T2 | T3 | T4 | T5 | T6 | |
| T1 | 1 | 1/3 | 1/3 | 1/3 | 1 | 1/3 | |
| T2 | 3 | 1 | 1 | 1 | 3 | 1 | |
| T3 | 3 | 1 | 1 | 1 | 3 | 1 | |
| T4 | 3 | 1 | 1 | 1 | 3 | 1 | |
| T5 | 1 | 1/3 | 1/3 | 1/3 | 1 | 1/3 | |
| T6 | 3 | 1 | 1 | 1 | 3 | 1 | |
| total | 14 | 14/3 | 14/3 | 14/3 | 14 | 14/3 | |

| Gender | | | | | | | |
|---|---|---|---|---|---|---|---|
| Teacher | T1 | T2 | T3 | T4 | T5 | T6 | |
| T1 | 1/14 | 1/14 | 1/14 | 1/14 | 1/14 | 1/14 | |
| T2 | 3/14 | 3/14 | 3/14 | 3/14 | 3/14 | 3/14 | |
| T3 | 3/14 | 3/14 | 3/14 | 3/14 | 3/14 | 3/14 | |
| T4 | 3/14 | 3/14 | 3/14 | 3/14 | 3/14 | 3/14 | |
| T5 | 1/14 | 1/14 | 1/14 | 1/14 | 1/14 | 1/14 | |
| T6 | 3/14 | 3/14 | 3/14 | 3/14 | 3/14 | 3/14 | |
| total | 14 | 14/3 | 14/3 | 14/3 | 14 | 14/3 | |

| Gender | | | | | | | |
|---|---|---|---|---|---|---|---|
| Teacher | T1 | T2 | T3 | T4 | T5 | T6 | Average |
| T1 | 0.072 | 0.072 | 0.072 | 0.072 | 0.072 | 0.072 | **0.072** |
| T2 | 0.214 | 0.214 | 0.214 | 0.214 | 0.214 | 0.214 | **0.214** |
| T3 | 0.214 | 0.214 | 0.214 | 0.214 | 0.214 | 0.214 | **0.214** |
| T4 | 0.214 | 0.214 | 0.214 | 0.214 | 0.214 | 0.214 | **0.214** |
| T5 | 0.072 | 0.072 | 0.072 | 0.072 | 0.072 | 0.072 | **0.072** |
| T6 | 0.214 | 0.214 | 0.214 | 0.214 | 0.214 | 0.214 | **0.214** |
| total | 14 | 14/3 | 14/3 | 14/3 | 14 | 14/3 | **1** |

First, the sum of each column is calculated. Then, each element of each column is divided by the corresponding sum. Finally, the preference vector is calculated such that each value corresponds to its row average.

Therefore, the preference vectors of the all criteria are shown in table 8.

Table 8: Preference vectors for the four criteria

| Teacher | Age | Gender | Load | Contract |
|---|---|---|---|---|
| T1 | **0.193** | **0.072** | **0.420** | **0.071** |
| T2 | **0.100** | **0.214** | **0.188** | **0.071** |
| T3 | **0.086** | **0.214** | **0.068** | **0.358** |
| T4 | **0.086** | **0.214** | **0.188** | **0.358** |
| T5 | **0.342** | **0.072** | **0.068** | **0.071** |
| T6 | **0.193** | **0.214** | **0.068** | **0.071** |

The third step of AHP is ranking the criteria to determine their relative importance or weights. Again, the pairwise comparison is used for this purpose.

Agreeing with the relative importance previously described, table 9 shows the criteria' weights which are calculated in a similar way as the preference vectors.

Table 9: Criteria's weights

| Criteria | Age | Gender | Load | Contract | **Weight** |
|---|---|---|---|---|---|
| Age | 1 | 3 | 1/5 | 1/3 | **0.122** |
| Gender | 1/3 | 1 | 1/7 | 1/5 | **0.057** |
| Load | 5 | 7 | 1 | 3 | **0.558** |
| Contract | 3 | 5 | 1/3 | 1 | **0.263** |
| Total | 28/3 | 16 | 176/105 | 68/15 | |

Now, the score of each teacher may be calculated as a vector/matrix multiplication of the weight of table 9 and the matrix of table 8. The obtained result is as follows:

- $S_1$ = Score T1 = $(0.193 \times 0.122) + (0.072 \times 0.057) + (0.420 \times 0.558) + (0.071 \times 0.263)$ = **0.281**
- $S_2$ = Score T2 = **0.148**
- $S_3$ = Score T3 = **0.158**
- $S_4$ = Score T4 = **0.222**
- $S_5$ = Score T5 = **0.102**
- $S_6$ = Score T6 = **0.089**

The set of score S = {$Si$, i = 1 ... 6} will be used in the next section to incorporate an overall satisfaction function to the GA in order to produce the requested time table satisfying most of the teachers' preferences.

Finally, we mention that AHP also incorporates a useful technique for checking the consistency of the decision evaluation. A consistency test may be applied to our study to verify if the considered pairwise comparisons are reliable. For further information about the consistency test, refer to [13].

## 5. COMBINING AHP AND GA (AHP/GA)

As it has been presented, the genetic algorithm proposes a fitness function used to test the feasibility of a produced solution. The classic fitness function of a GA applied to solve the TTP takes into consideration the time conflict. We define this function by $F_{conflict}$. Practically, $F_{conflict}$ may return the number of conflicts in the generated solution and a solution if feasible when the value of $F_{conflict}$ is 0 [15]. This task is implemented by several previous works.

In our (AHP/GA) approach, another function $F_{satisfaction}$ – *the satisfaction function*- is integrated to the GA fitness function, such that $F_{satisfaction} = \sum S_i \times M_i$. Where $S_i$, is the score of teacher $T_i$ generated by AHP, and $M_i$ is the number of sessions of teacher Ti that match its preferred schedule (see table 3). Observe that the maximum value of $F_{satisfaction}$ (called MaxF$_{satisfaction}$) is $\sum S_i \times L_i$, where $L_i$ is the load of Ti, and hence we obtain a perfect satisfaction.

As a result, we say that the solution of (AHP/GA) is feasible, when the two following conditions are respected:

1. $F_{conflict} = 0$
2. $F_{satisfaction} \geq$ ST ; where ST is a *satisfaction threshold* set by the TT designer. Note that, in order to obtain a perfect satisfaction, we only need to set ST to be $\sum S_i \times L_i$.

For our case study, MaxF$_{satisfaction}$ = (0.281×7) + (0.148×4) + (0.158×3) + (0.222×5) + (0.102×2) + (0.089×3) = 4.614.

Table 10 presents a randomly generated solution of AHP/GA with F$_{satisfaction}$ = (0.281×**5**) + (0.148×**2**) + (0.158×**1**) + (0.222×**5**) + (0.102×**2**) + (0.089×**2**) = 3.351, which is not feasible if we assume that ST = 4.

Table 10: A non-feasible solution of AHP/GA

| C1 | | | C2 | | |
|---|---|---|---|---|---|
| **D1** | **D2** | **D3** | **D1** | **D2** | **D3** |
| *T1* | *T1* | *T1* | *T4* | *T5* | *T6* |
| *T2* | *T2* | *T3* | *T1* | *T1* | *T1* |
| *T3* | *T4* | *T4* | *T1* | *T2* | *T2* |
| *T5* | *T6* | *T6* | *T3* | *T4* | *T4* |

Finally, note that a feasible solution with a perfect satisfaction may be generated by AHP/GA for our case study, and it is shown in table 11.

Table 11: Feasible solution of AHP/GA with perfect satisfaction

| C1 | | | C2 | | |
|---|---|---|---|---|---|
| **D1** | **D2** | **D3** | **D1** | **D2** | **D3** |
| *T1* | *T4* | *T1* | *T6* | *T2* | *T4* |
| *T6* | *T4* | *T3* | *T1* | *T2* | *T1* |
| *T2* | *T3* | *T6* | *T1* | *T1* | *T3* |
| *T2* | *T1* | *T5* | *T5* | *T4* | *T4* |

# 6. CONCLUSION

This paper presents a new approach that deals with the teachers' preferences while constructing the time table of a school schedule. Our approach consists of the integration of a satisfaction function to the genetic algorithm. The parameters of the satisfaction function are the teachers' loads and a set of scores calculated using the analytical hierarchy process. The key point of AHP in calculating the scores is the pairwise comparison of a set of teachers' criteria. The new approach is consequently a combination of AHP and GA, and it gives rise to a new methodology to solve the time table problem that we call AHP/GA.

In this paper, we present an informal description of our AHP/GA approach. The formal description and the implementation of the procedure are in perspective. In addition, we need to verify our contribution by applying the implemented AHP/GA to different case studies in Lebanese schools. Moreover, the consideration of more sophisticated constraints related to the courses order, and their integration into AHP/GA is another important perception. The declaration of a set of rules, for the manual interception while building the time table, is also suitable.

# Authors

**Ihab Sbeity**, the main author, is Associative Professor in Computer Science at the Faculty of Sciences in the Lebanese University. He received a *Maîtrise* in applied mathematics from the Lebanese university, a Master in computer science - systems and communications from Joseph Fourier University, France, and a PhD from INPG (*Institut National Polytechnique de Grenoble*), France. His PhD works are related to Performance Evaluation and System Design. Actually, his research interests include software performance engineering, UML modelling, and decision information system.

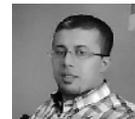

**Mohamed Dbouk**, is a full time Professor at the Lebanese University (Beirut-Lebanon), he coordinates a master-2 research degree "IDSS: Information and Decision Support Systems". He received his PhD from "Paris-Sud", France, 1997. His research interests include software engineering and information systems related issues, geographic information systems, data-warehousing and data-mining. 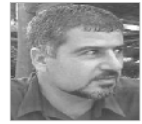

**Habib Kobeissi**, is Associate Professor of Data Structure at the Faculty of Sciences in the Lebanese University (since 1987). He holds a Bachelor of Science in Computer Science from Grenoble University, a Master of Science in Data Structure from the University of Paris VI, and a Ph.D. in Data Structure from the University of Paris V. Between 1979 and 1987, he worked as a Computer Scientist at several French companies. 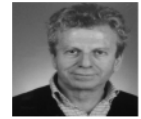